\documentclass[acmtog,nonacm]{acmart}
\acmSubmissionID{None}
\usepackage{booktabs} %
\usepackage{multirow}
\usepackage{tabularx}
\usepackage{wrapfig}
\usepackage{mathtools}
\usepackage{lipsum}
\usepackage{footmisc}

\usepackage[ruled]{algorithm2e} 

\SetAlFnt{\small}
\SetAlCapFnt{\small}
\SetAlCapNameFnt{\small}
\SetAlCapHSkip{0pt}

\acmJournal{TOG}

\begin{document}
\def \cf {{cf.\thinspace}}
\def \etal {{\emph{et al}.\thinspace}}
\def \eg {{\emph{e.g.}\thinspace}}
\def \ie {{\emph{i.e.}\thinspace}}
\def \vs {{\emph{v.s.}\thinspace}}
\def \etc {{etc.\ }}

\definecolor{myblue}{RGB}{0,211,255}
\definecolor{myorange}{RGB}{255,43,0}
\definecolor{myyellow}{RGB}{255,209,0}

\definecolor{LightCyan}{rgb}{0.88,1,1}
\definecolor{Gray}{gray}{0.9}


\newcommand{\name}{hierarchical neural semantic representation\xspace}
\newcommand{\abb}{HNSR\xspace}

\begin{teaserfigure}
  \centering
  \includegraphics[width=1\textwidth]{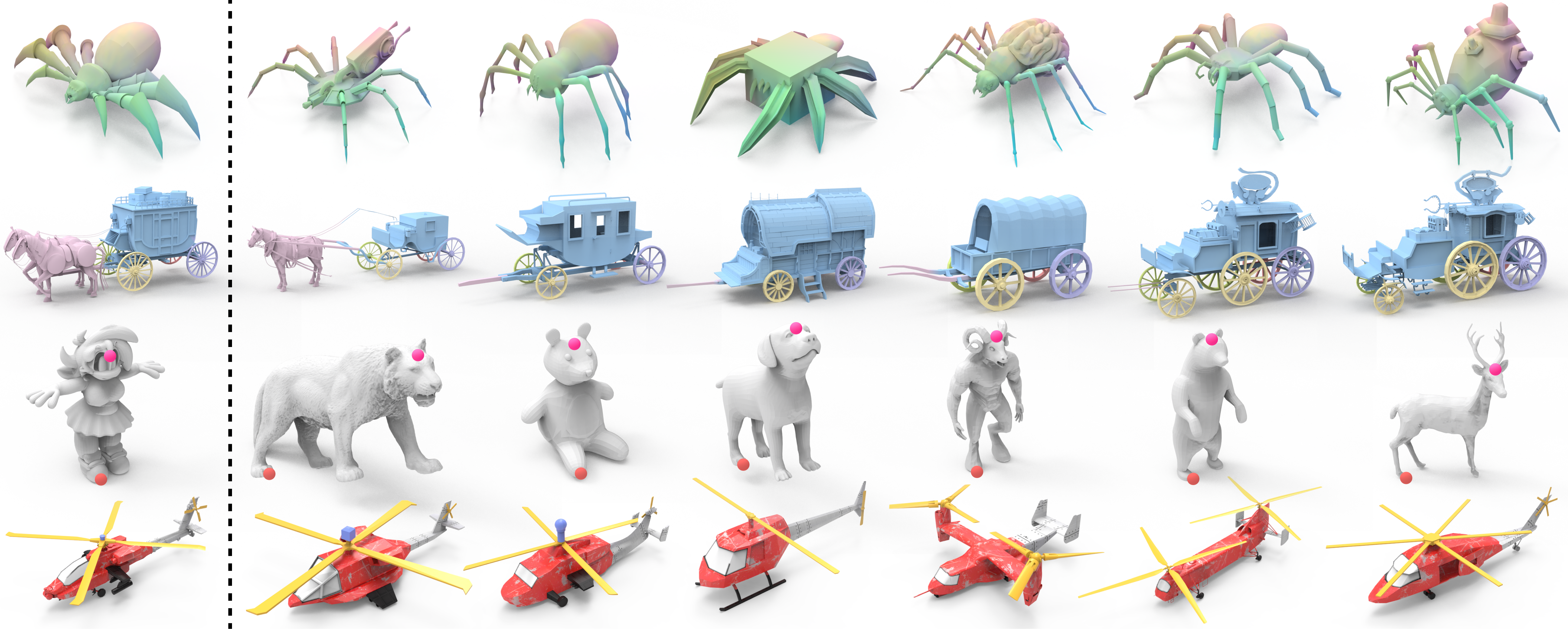}
  \caption{
  Our new framework is capable of predicting robust 3D correspondence between shapes of varying topological structures and geometric complexities,
  such as variations in the spider body and leg numbers (top row), the presence/absence of horses (2nd row), and variations in the main rotors of the helicopters (last row).
  Our framework also supports various applications: shape co-segmentation (2nd row), keypoint matching (3rd row), and texture transfer (last row). 
  }
  \label{fig:teaser}
  \vspace{2mm}
\end{teaserfigure}

\title{Hierarchical Neural Semantic Representation for 3D Semantic Correspondence}

\author{Keyu Du}
\affiliation{%
	\institution{University of Electronic Science and Technology of China} \country{China}}
\authornote{Both authors contributed equally to the paper.}
\author{Jingyu Hu}
\authornotemark[1]
\authornote{Corresponding author.}
\affiliation{%
	\institution{The Chinese University of Hong Kong} \country{HK SAR, China}}
\author{Haipeng Li}
\affiliation{%
	\institution{University of Electronic Science and Technology of China} \country{China}}
    \author{Hao Xu}
\affiliation{%
	\institution{The Chinese University of Hong Kong} \country{HK SAR, China}}
\author{Haibin Huang}
\affiliation{%
	\institution{TeleAI} \country{China}}

\author{Chi-Wing Fu}
\affiliation{%
	\institution{The Chinese University of Hong Kong} \country{HK SAR, China}}

\author{Shuaicheng Liu}
\affiliation{%
	\institution{University of Electronic Science and Technology of China} \country{China}}
\renewcommand\shortauthors{Du et al.}

\vspace{4mm}
\begin{abstract}

This paper presents a new approach to estimate accurate and robust 3D semantic correspondence with the hierarchical neural semantic representation.
Our work has three key contributions. 
%
%
%
First, we design the \name (\abb), which consists of a global semantic feature to capture high-level structure and multi-resolution local geometric features to preserve fine details, 
by carefully harnessing 3D priors from pre-trained 3D generative models.
Second, we design a progressive global-to-local matching strategy, which establishes coarse semantic correspondence using the global semantic feature, then iteratively refines it with local geometric features, yielding accurate and semantically-consistent mappings.
Third, our framework is training-free and broadly compatible with various pre-trained 3D generative backbones, demonstrating strong generalization across diverse shape categories.
Our method also supports various applications, such as shape co-segmentation, keypoint matching, and 
texture transfer, and generalizes well to structurally diverse shapes, with promising results even in cross-category scenarios. 
Both qualitative and quantitative evaluations show that our method outperforms previous state-of-the-art techniques.

\end{abstract}



\maketitle

\section{INTRODUCTION}
{
Establishing semantic correspondence between 3D shapes~\cite{van2011survey} has broad application values in computer graphics,~\eg, shape analysis~\cite{sidi2011unsupervised}, texture transfer~\cite{dinh2005texture}, and shape editing~\cite{sumner2004deformation, zhu2017deformation}.
However, accurate semantic-aware 3D correspondences are very challenging to achieve, as we need to build precise mappings between 3D shapes, while attentively considering the topological difference, object categories, and parts semantics, as well as the geometric details.
The task becomes even more challenging when handling shapes with missing parts and cross-category differences.
}

{
Traditional methods~\cite{ovsjanikov2012functional, huang2014functional} typically rely on hand-crafted shape descriptors to match shapes.
Hence, they often fall short of generalizing to handle diverse shapes.
Later,~\cite{litany2017deep, marin2020correspondence} propose to learn shape descriptors by using a neural network to detect shape correspondence.
However, these methods rely on annotated data, which is labor-intensive to create, and are often limited to specific domains, such as human bodies~\cite{melzi2019shrec} and animals~\cite{dyke2020shrec}. Such a narrow focus restricts their capabilities to handle a wider range of shapes with varying topologies and geometry, and thus their applicability in real scenarios.

}

{
Very recently, various 3D generative models~\cite{zhao2025hunyuan3d, zhang2024clay, hui2024make, li2025triposg} demonstrate rich semantic and geometric priors learned by pre-trained models when leveraging large-scale 3D shape data~\cite{deitke2023objaverse, deitke2023objaversexl}.
%
%
%
Though these models are designed mainly for shape generation, their latent representations implicitly capture rich geometric and semantic information. 
Hence, they have great potential values for downstream shape analysis,~\eg, 3D semantic correspondence and shape co-segmentation.
However, how to effectively exploit 3D generative models for shape analysis remains underexplored. 
}

{
In this work, we propose a new framework that harnesses pre-trained 3D generative models to construct a 
\textbf{H}ierarchical \textbf{N}eural \textbf{S}emantic \textbf{R}epre-sentation \textbf{(HNSR)}
for 3D semantic correspondence. 
Our representation consists of a global semantic feature that captures high-level structural information and a set of local geometric features at multiple resolutions that preserve fine-grained details. These multi-scale features are extracted from a pre-trained 3D generative model.
Building on it, we propose a progressive global-to-local matching strategy that first establishes coarse correspondences using a global semantic feature, then refines them with local geometric features.
This design effectively enables accurate and robust semantic correspondence between 3D shapes with structural and geometric variations. As the second row of Figure~\ref{fig:teaser} shows, our method achieves consistent shape co-segmentation results, even for shapes with complex topologies and fine geometries. The first and fourth rows present semantic correspondence and texture transfer results, manifesting that our method remains effective, despite missing parts between source and target shapes, demonstrating its strong robustness to structural inconsistency.
%
%

%
%
%
Our framework is compatible with a variety of pre-trained 3D generative models, including category-specific diffusion models~\cite{zheng2023locally, shim2023diffusion} and a 3D foundation model~\cite{xiang2024structured} trained on large-scale  datasets~\cite{deitke2023objaverse,deitke2023objaversexl}.
By exploiting the geometric and semantic priors learned by these models, our method achieves robust performance in diverse settings.
Notably, when combined with the 3D foundation model~\cite{xiang2024structured}, our method exhibits strong generalization to challenging cross-category scenarios. As the third row of Figure~\ref{fig:teaser} shows, our method is able to accurately match semantically corresponding positions such as the human foot and the lion’s paw. 
In particular, this is achieved in a training-free manner, as our method directly utilizes features extracted from the 3D generative backbones, {\em without\/} requiring additional supervision or task-specific fine-tuning.
}
{
Both quantitative and qualitative comparisons between our framework and various baselines show its superior performance.
Furthermore, we showcase that our framework can be applied in various downstream applications such as shape co-segmentation, keypoint matching, and texture transfer, highlighting its generalizability in real-world 3D shape analysis scenarios.

In summary, our contributions are summarized as follows.
\begin{itemize}
    \item We introduce a novel
    \textbf{H}ierarchical \textbf{N}eural \textbf{S}emantic \textbf{R}epre-sentation \textbf{(HNSR)}
    by extracting hierarchical features from pre-trained 3D generative models, enabling rich encoding of shape semantics and geometry.
    \item We propose a progressive global-to-local correspondence matching strategy that starts with coarse semantic matching and progressively refines it for precise correspondence.
    \item Our method requires no additional training and is compatible with various 3D generative backbones.
    We demonstrate the effectiveness of our approach through quantitative and qualitative experiments, achieving superior performance over state-of-the-art methods, while also supporting various downstream applications such as shape co-segmentation, keypoint matching, and texture transfer.

\end{itemize}
}
%

\section{RELATED WORK}
\subsection{3D Shape Correspondence}
%

3D shape correspondence~\cite{van2011survey, bronstein2008numerical, huang2008non, kim2011blended} is a fundamental task in computer graphics, aiming at establishing correspondences between 3D shapes. 
Among the existing works, the functional map emerges as a dominant approach. 
This approach computes shape descriptors to derive functional maps, which encode shape correspondences as compact representations within the eigenbasis of the Laplace-Beltrami operator (LBO). 
The traditional pipeline of functional-map-based methods~\cite{ren2018continuous, ovsjanikov2012functional, huang2014functional, nogneng2017informative, eisenberger2020smooth, rodola2014dense, litany2017fully, kovnatsky2015functional, ren2019structured} typically involves extracting handcrafted descriptors, followed by a functional-map estimation based on the descriptors.

With the advent of deep learning,~\cite{litany2017deep, zhuravlev2025denoising, cao2023self, halimi2019unsupervised, donati2020deep, roufosse2019unsupervised, sharma2020weakly} utilize the neural network to learn the shape descriptors from shape data in an end-to-end manner.
However, a key limitation persists: the reliance on spectral embeddings makes the correspondence quality highly dependent on the stability of the LBO eigenbasis.
When the input shapes exhibit significant 
 geometric or topological variations, the spectral decompositions can easily diverge, leading to misaligned eigenbases. 
This spectral instability propagates errors into the functional map, thereby degrading the correspondence accuracy.
Another line of works~\cite{groueix20183d, zheng2021deep, deng2021deformed, sumner2004deformation, zhu2017deformation, zhang2008deformation, eisenberger2019divergence} adopts a deformation-based paradigm, by deforming pre-defined templates to match the target geometry. 
While anchoring shapes to a common reference leads to a more consistent shape correspondence, relying on a fixed template limits the approach's flexibility, making it hard to handle shapes with large topological variations or fine-grained geometric details.
To overcome the limitations of existing approaches, we propose the first framework that harnesses the power of pre-trained 3D generative models to establish high-quality 3D semantic correspondence, enabling us to handle shapes with diverse topologies, fine-grained geometric details, and even cross-category variations.
In particular, our approach operates in a training-free manner. 
It generalizes well for different shape classes, even without labeled correspondence data and fine-tuning; see Section~\ref{ssec:visual_res}.

\vspace{-2mm}
\subsection{Semantic Correspondence}
%
%
%
%

Semantic correspondence is a relatively new topic, focusing on establishing a meaningful mapping between different instances by locating and matching semantically similar positions.
Unlike traditional correspondence methods~\cite{lowe2004distinctive, bay2006surf}, which focus on local appearance or geometric similarity, semantic correspondence targets higher-level understanding, aiming at a more robust matching even with variations in geometries, structures, and object categories.
Recent works study semantic correspondence mainly on images,~\eg,~\cite{rocco2018end, jiang2021cotr, jin2021image, lee2021patchmatch}.
%
Though these methods demonstrate promising results on moderate appearance and pose variations, their performance depends heavily on costly annotated datasets, thus limiting their applicability.
More recent methods~\cite{tang2023emergent, zhang2021datasetgan, peebles2022gan, luo2023diffusion, zhang2023tale, min2020learning} explore unsupervised techniques by making use of pre-trained image generative models~\cite{goodfellow2014generative, ho2020denoising}.
These approaches often utilize spatially-aligned representations and identify semantically-consistent regions based on feature similarity to enable correspondence estimation.
Research on 3D semantic correspondence is largely unexplored, with only a few existing works.
\cite{abdelreheem2023zero} propose a two-stage approach that first segments 3D shapes then establishes a mapping between segments to yield a coarse-level semantic correspondence. 
Yet, the method relies heavily on the quality of the segmentation, so the resulting semantic correspondence is often coarse and imprecise.
~\cite{zhu2024densematcher, dutt2024diffusion, liu2025stable, sun2024srif}
establish 3D semantic correspondences by projecting 2D features onto 3D shapes and aggregating the features across views, whereas~\cite{liu2025stable, sun2024srif} employ shape registration to construct correspondences.
However, as image features lack explicit 3D geometric information, directly transferring them to 3D often leads to noisy and inconsistent predictions.
The above limitations underscore the need for dedicated 3D descriptors that capture both fine-grained geometry and high-level semantics for accurate 3D semantic correspondence.
In this work, we design a new approach that operates directly in 3D.
We build our representation and perform the correspondence matching {\em entirely in the 3D space\/}, bypassing the need to pre-segment the shapes.
\vspace{-2mm}
\subsection{3D Generative Modeling}

Recent advances in large-scale 3D datasets such as Objaverse~\cite{deitke2023objaverse} and Objaverse-XL~\cite{deitke2023objaversexl} have paved the way for the development of 3D foundation models~\cite{li2025triposg, zhao2025hunyuan3d, hui2024make, zhang2024clay, xiang2024structured, hong20243dtopia}. 
These models are capable of generating high-quality and detailed 3D shapes, while generalizing effectively to unseen shapes of complex topologies and intricate details.
Such strong generalizability demonstrates the robust shape priors acquired by the models during training.
Yet, little has been done on how to effectively harness these powerful geometric priors for downstream shape analysis tasks such as shape co-segmentation and correspondence.

\section{METHOD}
\subsection{OVERVIEW}

\begin{figure*}
    \centering
    \includegraphics[width=1\linewidth]{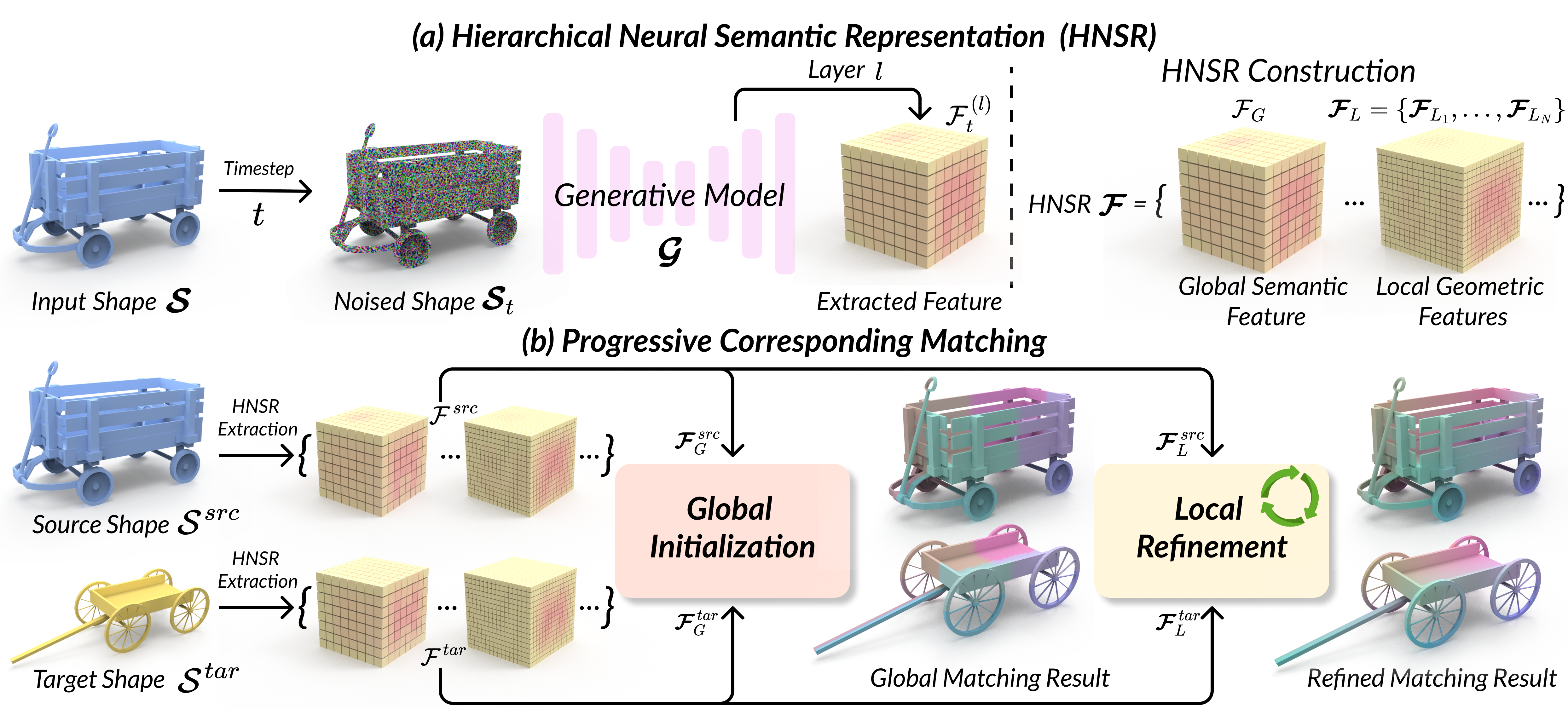}
    \caption{
    Overview of our approach.
    (a) We design the \name (\abb) using multi-scale features extracted from a pre-trained 3D generative model $\boldsymbol{\mathcal{G}}$.
    Given an input shape $\boldsymbol{\mathcal{S}}$, we add noise to obtain $\boldsymbol{\mathcal{S}_t}$ at time step $t$, and extract features $\boldsymbol{\mathcal{F}}^{(l)}_t$ from the $l$-th layer of $\boldsymbol{\mathcal{G}}$.
    By aggregating a global feature and fine features across multiple layers, we construct the \abb $\boldsymbol{\mathcal{F}}$ for the input shape. 
    (b) Given a pair of source and target shapes, we propose a progressive global-to-local matching strategy for 3D semantic correspondence.
    We first construct their respective \abb representations, extracting both global semantic features and local geometric features.
    The global features are used to perform coarse correspondence initialization, which is then iteratively refined using the local features at higher resolutions, resulting in accurate and semantically consistent correspondences.
    }
    \label{fig: overview}
    \vspace{-4mm}
\end{figure*}

Given a pair of source $\boldsymbol{\mathcal{S}}^{src}$ and target $\boldsymbol{\mathcal{S}}^{tar}$ shapes, our goal is to establish the 3D semantic correspondence between them.
As a prerequisite, both shapes are represented in a voxelized form, which serves as the input to our framework.
Figure~\ref{fig: overview} provides an overview of our two-stage framework:
\begin{itemize}
    \vspace{1mm}
    \item \textbf{Hierarchical Neural Semantic Representation.} First, we design the \name (\abb), denoted as $\boldsymbol{\mathcal{F}}$, to encode both semantic and geometric information of a 3D shape in a multi-resolution manner. Given input shape $\boldsymbol{\mathcal{S}}$ in voxel form, we extract $\boldsymbol{\mathcal{F}}$ from a pre-trained 3D generative model $\boldsymbol{\mathcal{G}}$. Specifically, we perturb $\boldsymbol{\mathcal{S}}$ and pass it through $\boldsymbol{\mathcal{G}}$ to extract intermediate features from selected network layers. We then organize these features into global semantic feature $\boldsymbol{\mathcal{F}}_G$ and a set of local geometric features $\boldsymbol{\mathcal{F}}_L$, forming the \abb: $\boldsymbol{\mathcal{F}} = (\boldsymbol{\mathcal{F}}_G, \boldsymbol{\mathcal{F}}_L)$.

    \vspace{1mm}
    \item \textbf{Progressive Correspondence Matching.}  
    Second, we design a progressive global-to-local matching strategy to establish accurate semantic correspondences between 3D shapes. Given the extracted 
    \abb of the source and target shapes, our method proceeds in two stages: (i) Global Initialization, which computes coarse semantic correspondences using the low-resolution global feature to localize semantically similar regions; and (ii) Local Refinement, which iteratively refines the correspondences using multi-scale local geometric features, progressively narrowing down the matching region at each level. 
    This strategy encourages semantically meaningful and spatially precise correspondences.
\end{itemize}

\begin{teaserfigure}
  \centering
  \includegraphics[width=1\textwidth]{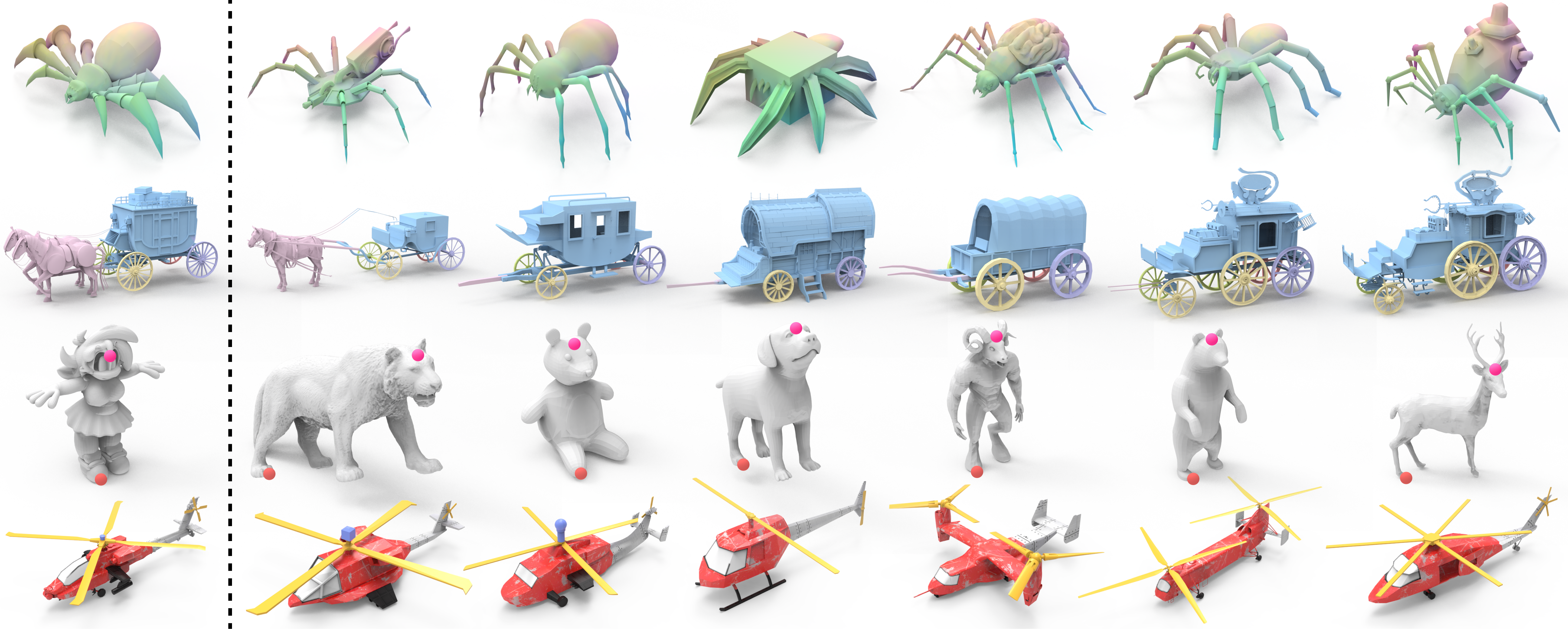}
  \caption{
  Our new framework is capable of predicting robust 3D correspondence between shapes of varying topological structures and geometric complexities,
  such as variations in the spider body and leg numbers (top row), the presence/absence of horses (2nd row), and variations in the main rotors of the helicopters (last row).
  Our framework also supports various applications: shape co-segmentation (2nd row), keypoint matching (3rd row), and texture transfer (last row). 
  }
  \label{fig:teaser}
  \vspace{2mm}
\end{teaserfigure}

\subsection{Hierarchical Neural Semantic Representation}
\label{ssec3: nsr_extraction}

Given an input 3D shape $\boldsymbol{\mathcal{S}}$ in a voxel-based representation, our goal is to build the \textbf{Hierarchical Neural Semantic Representation (\abb) $\boldsymbol{\mathcal{F}}$}, composed of a global semantic feature and a set of local geometric features at multiple resolutions.
These multi-scale features are hierarchically extracted from a pre-trained 3D generative model $\boldsymbol{\mathcal{G}}$ and are designed to capture both high-level semantic structures and fine-grained geometric details.
We construct the \abb $\boldsymbol{\mathcal{F}}$ in two stages: (i) \emph{Feature Extraction} and (ii) \emph{\abb Construction}, as Figure~\ref{fig: overview}(a) illustrates.
The specific voxel format of $\boldsymbol{\mathcal{S}}$ (\eg, SDF, occupancy, or latent code) depends on the backbone of the generative model $\boldsymbol{\mathcal{G}}$.
The details of both the input format and the corresponding generative backbones will be provided in Section~\ref{ssec:backbone}.

\vspace*{1mm}
\noindent\textbf{Feature Extraction.}
To faithfully leverage the 3D priors encoded in the generative model $\boldsymbol{\mathcal{G}}$, we follow a feature extraction strategy that mirrors the model’s training. In particular, for diffusion-based generative models, training involves denoising shapes corrupted by increasing levels of Gaussian noise. To simulate this, we perturb the input shape $\boldsymbol{\mathcal{S}}$ to obtain a noisy version $\boldsymbol{\mathcal{S}}_t$ at diffusion timestep $t$:
\begin{equation}
\boldsymbol{\mathcal{S}}_t = \sqrt{\bar{\alpha}_t} \boldsymbol{\mathcal{S}} + \sqrt{1 - \bar{\alpha}_t} \boldsymbol{\epsilon}, \quad \boldsymbol{\epsilon} \sim \mathcal{N}(\mathbf{0}, \mathbf{I}),
\end{equation}
where $\bar{\alpha}_t$ is a predefined noise schedule on 
the noise level at timestep $t \in [1, T]$. 
We then feed the noisy shape $\boldsymbol{\mathcal{S}}_t$ into the generative model $\boldsymbol{\mathcal{G}}$ and extract intermediate features from the selected layer $l$:
\begin{equation}
\boldsymbol{\mathcal{F}_t^{(l)}} = \boldsymbol{\mathcal{G}}^{(l)}(\boldsymbol{\mathcal{S}}_t),
\label{eq2}
\end{equation}
where $\boldsymbol{\mathcal{G}}^{(l)}$ denotes the output of the $l$-th layer. Each extracted feature $\boldsymbol{\mathcal{F}}_t^{(l)}$ encodes the model’s internal representation of the shape, influenced by the layer depth $l$.
By sampling across network layers, we obtain the hierarchical representations that capture the shape's structure across varying semantic and geometric scales.

\vspace{1mm}
\noindent\textbf{\abb Construction.}
To build a representation that incorporates both coarse semantics and fine geometric cues, we organize the extracted features into a hierarchical structure based on their associated network layers.
This design is motivated by~\cite{pan2023drag}: deeper layers (farther from output) in neural networks tend to encode global semantics, whereas shallower layers (closer to the output) focus more on local geometries. 
Therefore, we define the \abb $\boldsymbol{\mathcal{F}}$ as comprising two components:
\begin{itemize}
    \item \textbf{Global semantic feature} $\boldsymbol{\mathcal{F}}_G$: a single feature extracted from a deep layer, capturing high-level semantic structure and contextual information of the shape; and
    \vspace{1mm}
    \item \textbf{Local geometric features} $\boldsymbol{\mathcal{F}}_L=\{\boldsymbol{\mathcal{F}}_{L_{1}},...,\boldsymbol{\mathcal{F}}_{L_{N}}\}$: a set of features collected from multiple shallow layers, each encoding fine-grained geometric and topological details at different levels of resolution.
\end{itemize}
This \name, $\boldsymbol{\mathcal{F}} = (\boldsymbol{\mathcal{F}}_G, \boldsymbol{\mathcal{F}}_L)$, yields a multi-scale representation that is both semantically meaningful and geometrically precise. 
The resulting \abb $\boldsymbol{\mathcal{F}}$ is training-free, serving as an efficient and versatile representation that can be applied across various pre-trained 3D generative models. 

\vspace{-2mm}
\subsection{Progressive Correspondence Matching}

\begin{figure}[t]
    \centering
    \includegraphics[width=0.95\linewidth]{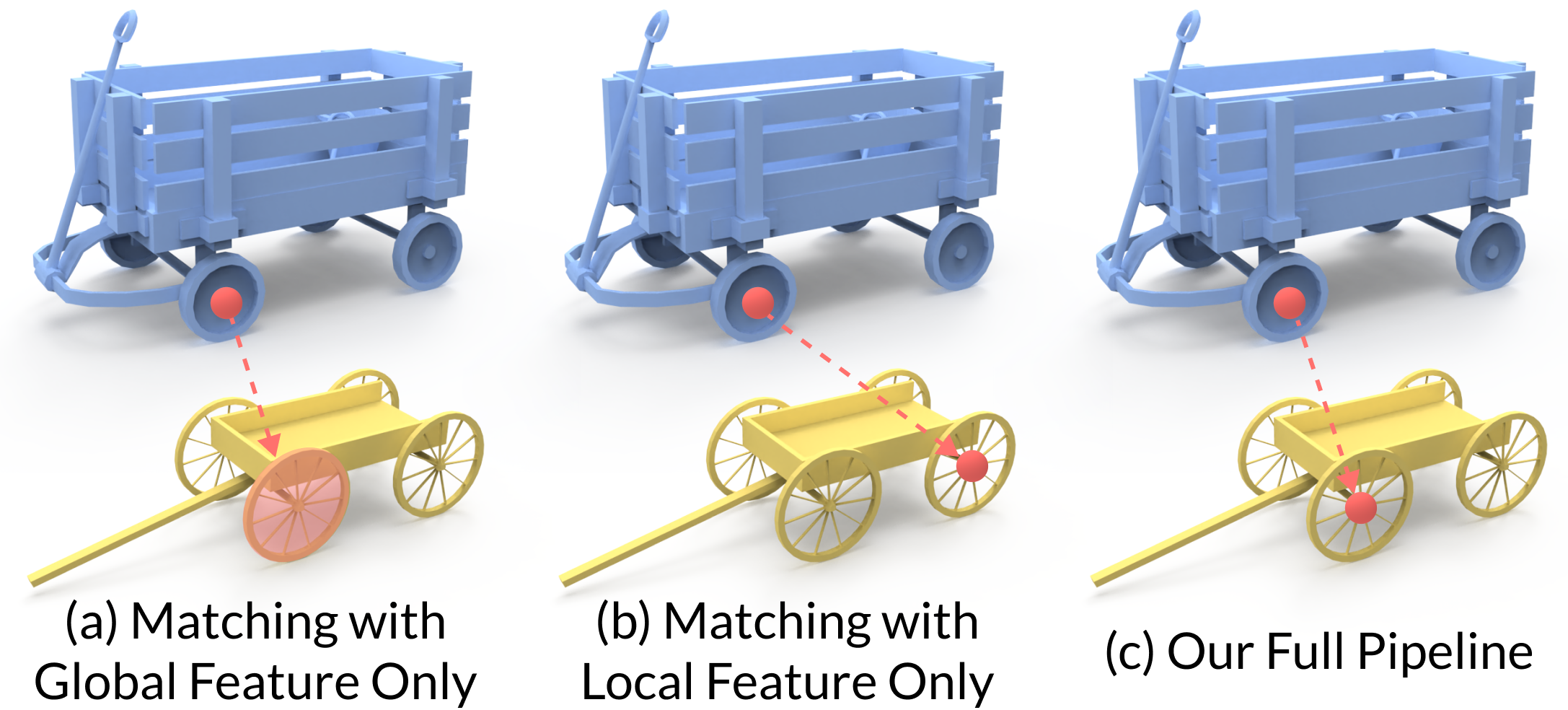}
    \vspace{-3mm}
    \caption{
    Visual comparison of shape correspondence with global and local features. 
    (a) Using global semantic features alone yields coarse alignments due to limited spatial resolution.
    (b) Using local geometric features alone may introduce mismatches in regions with similar local geometry but different global semantics.
    (c) Our progressive global-to-local matching scheme combines both, achieving correspondences that are semantically consistent and geometrically precise.
    }
    \vspace{-4mm}
    \label{fig: sole_fea_issue}
\end{figure}

Given a source shape $\boldsymbol{\mathcal{S}}^{src}$ and a target shape $\boldsymbol{\mathcal{S}}^{tar}$, we first extract their respective \abb using the strategy described in Section~\ref{ssec3: nsr_extraction}: $\boldsymbol{\mathcal{F}}^{src} = (\boldsymbol{\mathcal{F}}_G^{src}, \boldsymbol{\mathcal{F}}_L^{src})$ and $\boldsymbol{\mathcal{F}}^{tar} = (\boldsymbol{\mathcal{F}}_G^{tar}, \boldsymbol{\mathcal{F}}_L^{tar})$.
A natural idea is to directly match features using either the global or local features.
However, relying solely on the global semantic features often leads to low-resolution and imprecise correspondences, as Figure~\ref{fig: overview}(b) (global matching result) and Figure~\ref{fig: sole_fea_issue}(a) illustrate.
On the other hand, matching only the local geometric features can lead to incorrect results in geometrically similar regions with different semantic meanings,~\eg, confusing the front and back wheels of a car due to their local appearance similarity, as Figure~\ref{fig: sole_fea_issue}(b) illustrates.

To mitigate this issue, we introduce a progressive global-to-local matching scheme. 
This scheme first leverages global semantic features to establish coarse correspondences that reflect high-level semantic structural relationships, and then incrementally refines them using multi-scale local geometric features to obtain spatially-accurate correspondence, as shown in Figure~\ref{fig: sole_fea_issue}(c).
The overall matching process consists of two stages: (i) \emph{Global Initialization}, which computes global coarse correspondences using $\boldsymbol{\mathcal{F}}_G^{src}$ and $\boldsymbol{\mathcal{F}}_G^{tar}$; and (ii) \emph{Local Refinement}, which progressively improves the correspondence using the local features $\boldsymbol{\mathcal{F}}_L^{src}$ and $\boldsymbol{\mathcal{F}}_L^{tar}$.

\noindent\textbf{Global Initialization.}
We begin by computing a coarse semantic alignment using the global semantic features $\boldsymbol{\mathcal{F}}_G^{src}$ and $\boldsymbol{\mathcal{F}}_G^{tar}$. 
For each voxel position $\mathbf{v}^{src}$ in $\boldsymbol{\mathcal{S}}^{src}$, we aim to identify a coarse matching region $\boldsymbol{\mathcal{R}}_0^{tar} \subset \boldsymbol{\mathcal{S}}^{tar}$ in the target shape that captures semantically similar content.
Specifically, given the source position $\mathbf{v}^{src}$, we identify its corresponding position $\mathbf{v}^{src}_G$ in the source global feature grid $\boldsymbol{\mathcal{F}}_G^{src}$, which typically has lower spatial resolution than $\boldsymbol{\mathcal{S}}^{src}$ due to downsampling in the generative model.
This mapping is computed by dividing the voxel coordinates by the resolution scale factor, depending on the feature resolution.
To find a semantically similar region in the target shape, we compare the global feature at position $\mathbf{v}^{src}_G$ with all feature vectors in the target global feature grid $\boldsymbol{\mathcal{F}}_G^{tar}$ based on cosine similarity. 
Specifically, for each spatial position $\mathbf{v}$ in the target global feature grid $\boldsymbol{\mathcal{F}}_G^{tar}$, we compute the cosine similarity between the source feature vector $\boldsymbol{\mathcal{F}}_G^{src}[\mathbf{v}^{src}_G]$ and the target feature vector $\boldsymbol{\mathcal{F}}_G^{tar}[\mathbf{v}]$. 
The most semantically similar position $\mathbf{v}_G^{tar}$ is then selected as follows:
\begin{equation}
\mathbf{v}_G^{tar} = \arg\min_{\mathbf{v} \in \boldsymbol{\mathcal{F}}_G^{tar}} \delta_{\cos}\left(\boldsymbol{\mathcal{F}}_G^{src}[\mathbf{v}^{src}_G], \boldsymbol{\mathcal{F}}_G^{tar}[\mathbf{v}]\right),
\end{equation}
where $\delta_{\cos}(\cdot, \cdot)$ denotes cosine similarity. 
Since the global feature grid $\boldsymbol{\mathcal{F}}_G^{tar}$ is defined at a lower resolution than the voxel grid $\boldsymbol{\mathcal{S}}^{tar}$, the matched position $\mathbf{v}_G^{tar}$ corresponds to a coarse region in $\boldsymbol{\mathcal{S}}^{tar}$.
Therefore, we map $\mathbf{v}_G^{tar}$ back to the coarse region $\boldsymbol{\mathcal{R}}_0^{tar} \subset \boldsymbol{\mathcal{S}}^{tar}$ by upsampling.
This establishes a global mapping, where each voxel position $\mathbf{v}^{src}$ in $\boldsymbol{\mathcal{S}}^{src}$ is associated with a corresponding region $\boldsymbol{\mathcal{R}}_0^{tar}$ in the target shape.
This region serves as a semantic prior to guide the subsequent local refinement process.

\noindent\textbf{Local Refinement.}
In this stage, given a source voxel position $\mathbf{v}^{src}$ in $ \boldsymbol{\mathcal{S}}^{src}$ and its corresponding coarse semantic region $\boldsymbol{\mathcal{R}}_0^{tar} \subset \boldsymbol{\mathcal{S}}^{tar}$ obtained from the global initialization stage, our goal is to progressively refine this region using the local geometric features. 
Specifically, we utilize the local features $\boldsymbol{\mathcal{F}}_L = \{\boldsymbol{\mathcal{F}}_{L_1}, \dots, \boldsymbol{\mathcal{F}}_{L_N}\}$ to obtain a sequence of increasingly precise regions $\{ \boldsymbol{\mathcal{R}}_i^{tar}|\boldsymbol{\mathcal{R}}_i^{tar} \subset \boldsymbol{\mathcal{R}}_{i-1}^{tar} \subset\ \boldsymbol{\mathcal{S}^{tar}} $ and $ i\in[1, N]\}$.
This progressive refinement ultimately yields an accurate corresponding position $\mathbf{v}^{tar}$ in $\mathcal{S}^{tar}$.
At each refinement level $i$, we first downsample the voxel coordinates in $\boldsymbol{\mathcal{R}}_{i-1}^{tar}$ to match the resolution of the target local feature $\boldsymbol{\mathcal{F}}_{L_i}^{tar}$.
This produces a restricted index set $\mathcal{A}_{i}^{tar}$ within the spatial domain of $\boldsymbol{\mathcal{F}}_{L_i}^{tar}$, which serves as the candidate search region at level $i$.
For the voxel position $\mathbf{v}^{src} $ in $\boldsymbol{\mathcal{S}}^{src}$, we downsample its coordinate to obtain $\mathbf{v}_{i}^{src}$ in the source’s local feature grid $\boldsymbol{\mathcal{F}}_{L_i}^{src}$ and extract its corresponding feature vector $\boldsymbol{\mathcal{F}}_{L_i}^{src}[\mathbf{v}_{i}^{src}]$. 
We then identify the best matching position in the restricted target region according to the cosine similarity:

\vspace{-2mm}
\begin{equation}
\mathbf{v}_{i}^{tar} = \arg\min_{\mathbf{v} \in \mathcal{A}_{i}^{tar}} \delta_{\cos} \left( \boldsymbol{\mathcal{F}}_{L_i}^{src}[\mathbf{v}_{i}^{src}], \boldsymbol{\mathcal{F}}_{L_i}^{tar}[\mathbf{v}] \right).
\end{equation}
\vspace{-3mm}

\noindent
where position $\mathbf{v}_{i}^{tar}$ corresponds to a voxel in $\boldsymbol{\mathcal{F}}_{L_i}^{tar}$, from which we upsample back to $\mathcal{S}^{tar}$ to define a finer region $\boldsymbol{\mathcal{R}}_{i}^{tar} \subset \boldsymbol{\mathcal{R}}_{i-1}^{tar} \subset \boldsymbol{\mathcal{S}}^{tar}$. 
This region is then used as input for the next refinement level. 
Repeating this process over all levels $N$ times enables us to obtain progressively refined correspondence positions, ultimately leading to more accurate correspondence between the source and target shapes.
While local refinement depends on global initialization, the global stage typically provides rough but robust correspondences, avoiding failures from bad global initialization.

\subsection{Backbones for Neural Semantic Representation}
\label{ssec:backbone}

Our neural semantic representation extraction framework is designed to work with diverse pre-trained 3D generative models, even though the form of the input shape $\boldsymbol{\mathcal{S}}$ and the structure of the generator $\boldsymbol{\mathcal{G}}$ vary across different backbones.
In this section, we describe three representative backbones that can be incorporated in our framework and clarify how $\boldsymbol{\mathcal{S}}$ and $\boldsymbol{\mathcal{G}}$ are defined in each case.

\vspace*{1mm}
\noindent\textbf{SDF-Diffusion}~\cite{shim2023diffusion}:
The input shape $\boldsymbol{\mathcal{S}}$ is represented as a voxelized signed distance field (SDF) volume at a resolution of $32^3$, where each voxel encodes the signed distance to the shape surface.
The generative model $\boldsymbol{\mathcal{G}}$ is a U-Net-based diffusion model trained to synthesize such SDF volumes via iterative denoising.
While SDF-Diffusion further employs a separate super-resolution module to upsample the generated SDFs into high-resolution outputs, this module is not involved in our NSR extraction process.
\vspace*{1mm}
\noindent\textbf{LAS-Diffusion}~\cite{zheng2023locally}:  
$\boldsymbol{\mathcal{S}}$ is a $64$$\times$$64$$\times$$64$ binary occupancy grid.
The generator $\boldsymbol{\mathcal{G}}$ is a U-Net-based diffusion model trained to generate shapes represented as occupancy grids via iterative denoising, which forms the first stage of the LAS-Diffusion pipeline. The second stage is a separate super-resolution module that refines the coarse shape into a high-quality signed distance field (SDF), but it is not involved in our NSR extraction.
%
%

%
\vspace*{1mm}
\noindent\textbf{TRELLIS}~\cite{xiang2024structured}:  
$\boldsymbol{\mathcal{S}}$ is a voxel structural embedding at a resolution of $16^3$, representing the geometry of the input shape.
TRELLIS utilizes multi-view image features to construct a structured latent representation ({\sc{SLat}}) through 
a two-stage rectified flow process: 
first predicting a voxel structure embedding $\boldsymbol{\mathcal{S}}$, capturing geometry structure, then generating fine-grained latent codes.
The generator $\boldsymbol{\mathcal{G}}$ refers to the first stage 
that produces $\boldsymbol{\mathcal{S}}$. 
Specifically, we adopt the stage-one voxel embeddings, which provide a volumetric representation naturally suited to our voxel-based pipeline, whereas the stage-two sparse codes are less compatible.

\begin{table}[t]
    \centering
    \caption{
    Accuracy and error comparison for shape correspondence. 
    We report the accuracy within a 1\% error tolerance, comparing our method with previous works.
    ``--'' denotes unavailable results.
    The best and second-best results are highlighted in \textbf{bold} and \underline{underlined}. 
    Since 3D-CODED~\cite{groueix20183d} does not offer a pre-trained model for the SHREC' 20 dataset, a comparison with this method is not provided in this dataset.
    }
    \label{tab: pointwise}
    \setlength\tabcolsep{5pt}
    \resizebox{1\linewidth}{!}{
        \small
        \begin{tabular}{l|cc|cc}
        \toprule
        \multirow{2}{*}{Methods} & \multicolumn{2}{c|}{SHREC' 19} & \multicolumn{2}{c}{SHREC' 20} \\
        \cmidrule{2-5}
        & \textit{Accuracy} $\uparrow$ & \textit{Error} $\downarrow$ & \textit{Accuracy} $\uparrow$ & \textit{Error} $\downarrow$ \\
        \midrule
        DPC & 17.4 & 6.3 & 31.1 & 2.1  \\
        SE-ORNet & 21.4 & 4.6 & 31.7 & 1.0  \\
        3D-CODED & 2.1 & 8.1 & -- &  --  \\
        FM + WKS & 4.4 & 3.3 & 4.1 & 7.3  \\
        DenseMatcher & 11.3 & 6.5 & 27.8 & 2.7 \\
        DIFF3F & \underline{26.4} & 1.7 & \underline{72.6} & 0.9 \\
        DIFF3F + FM & 21.6 & \underline{1.5} & 62.3 & \underline{0.7} \\

        \midrule
        HSNR (Ours) &\textbf{37.4} & \textbf{1.3} & \textbf{83.8} & \textbf{0.5}  \\
        \bottomrule
    \end{tabular}}
    \vspace{-3mm}
\end{table}
\begin{table}[t]
    \centering
    \caption{
    Comparing geodesic error
    and edge distortion ratio for shape correspondence.
    Notice that our method consistently achieves the best results (lowest geodesic error and edge distortion ratio closer to one) for all cases. 
    The units of the geodesic error is $10^{-3}$.}
    \vspace{-3mm}
    \label{tab: geo_smooth}
    \setlength\tabcolsep{6pt}
    \resizebox{1\linewidth}{!}{
        \small
        \begin{tabular}{l|cc|cc}
        \toprule
        \multirow{2}{*}{Methods} & \multicolumn{2}{c|}{SHREC' 19} & \multicolumn{2}{c}{SHREC' 20} \\
        \cmidrule{2-5}
        & \textit{Error} $\downarrow$ & \textit{Ratio} $\sim$ 1   & \textit{Error} $\downarrow$ & \textit{Ratio} $\sim$  1 \\
        \midrule
        DenseMatcher & 96 & 1.62 & 33 & 1.43 \\
        DIFF3F & 21 & 1.36 & 11 & 1.14 \\

        \midrule
        HSNR (Ours) &\textbf{15} & \textbf{1.24} & \textbf{6} & \textbf{1.11}  \\
        \bottomrule
    \end{tabular}}
    \vspace{-3mm}
\end{table}

\begin{table*}[th]
    \centering
    \caption{
    Per-part IoU comparison on the co-segmentation task. 
    The best and second-best results are marked in \textbf{bold} and \underline{underlined}.
    Our method with all three backbones consistently outperforms the state-of-the-art method across all categories, demonstrating the effectiveness and robustness of our framework.}
    \vspace{-3mm}
    \label{tab: Co_seg_IOU}
    \setlength{\tabcolsep}{5pt}
    \resizebox{\linewidth}{!}{
    \begin{tabular}{c|c|c|c|c|c|c|c|c|c|c|c|c|c|c}
        \toprule
        Methods  & \texttt{plane} & \texttt{cap} & \texttt{chair} & \texttt{earphone} & \texttt{guitar} & \texttt{knife} & \texttt{lamp} & \texttt{motorbike} & \texttt{mug} & \texttt{pistol} & \texttt{rocket} & \texttt{skateboard} & \texttt{table} & average\\
        \midrule
        BAE-Net & 67.8 & 82.1 & 66.5 & 52.1 & 53.9 & 41.5 & 75.3 & 23.1 & 95.3 & 31.4 & 42.1 & 65.7 & 76.5 & 60.7  \\
        RIM-Net & 61.9 & 59.3 & 75.7 & 70.8 & 46.3 & 40.7 & 60.6 & 28.7 & 67.6 & 36.9 & 36.7 & 65.9 & 71.4 & 53.1  \\
        SATR  & 40.6 & 22.9 & 58.3 & 22.6 & 50.6 & 53.7 & 47.6 & 16.6 & 54.3 & 23.3 & 36.7 & 48.9 & 37.5 & 42.1\\
        DAE-Net & 76.3 & 82.4 & 82.3 & 76.9 & 88.1 & 82.1 & 73.2 & 48.4 & 95.6 & 73.2 & 38.7 & 69.2 & 74.5 & 74.7\\
        DenseMatcher & 52.1 & 42.5 & 63.2 & 35.9 & 54.1 & 41.9 & 62.3 & 19.8 & 68.4 & 20.8 & 41.2 & 58.0 & 63.2 & 48.2\\
        HNSR + SDF-Diffusion  & 63.1 & 80.4 & 71.7 & \textbf{80.7} & 89.2 & 79.5 & 77.3 & \underline{56.7} & \textbf{96.7} & \underline{83.7} & 46.4 & 80.9 & 70.0 & 75.8\\
        HNSR + LAS-Diffusion  & \textbf{79.4} & \textbf{84.1} & \textbf{86.0} & \underline{80.6} & \underline{90.3} & \underline{83.9} & \textbf{83.8} & \textbf{62.6} & \underline{96.5} & 83.2 & \textbf{60.2} & \textbf{81.9} & \underline{88.3} & \textbf{82.7}\\
        HNSR + TRELLIS  & \underline{77.5} & \underline{83.6} & \underline{83.1} & 77.4 & \textbf{90.6} & \textbf{85.8} & \underline{81.2} & 51.2 & 95.1 & \textbf{85.7} & \underline{57.8} & \underline{81.2} & \textbf{89.2} & \underline{81.0}\\
        \bottomrule
    \end{tabular}}
    \vspace{-3mm}
\end{table*}

\begin{figure}[t]
    \centering
    \includegraphics[width=0.99\linewidth]{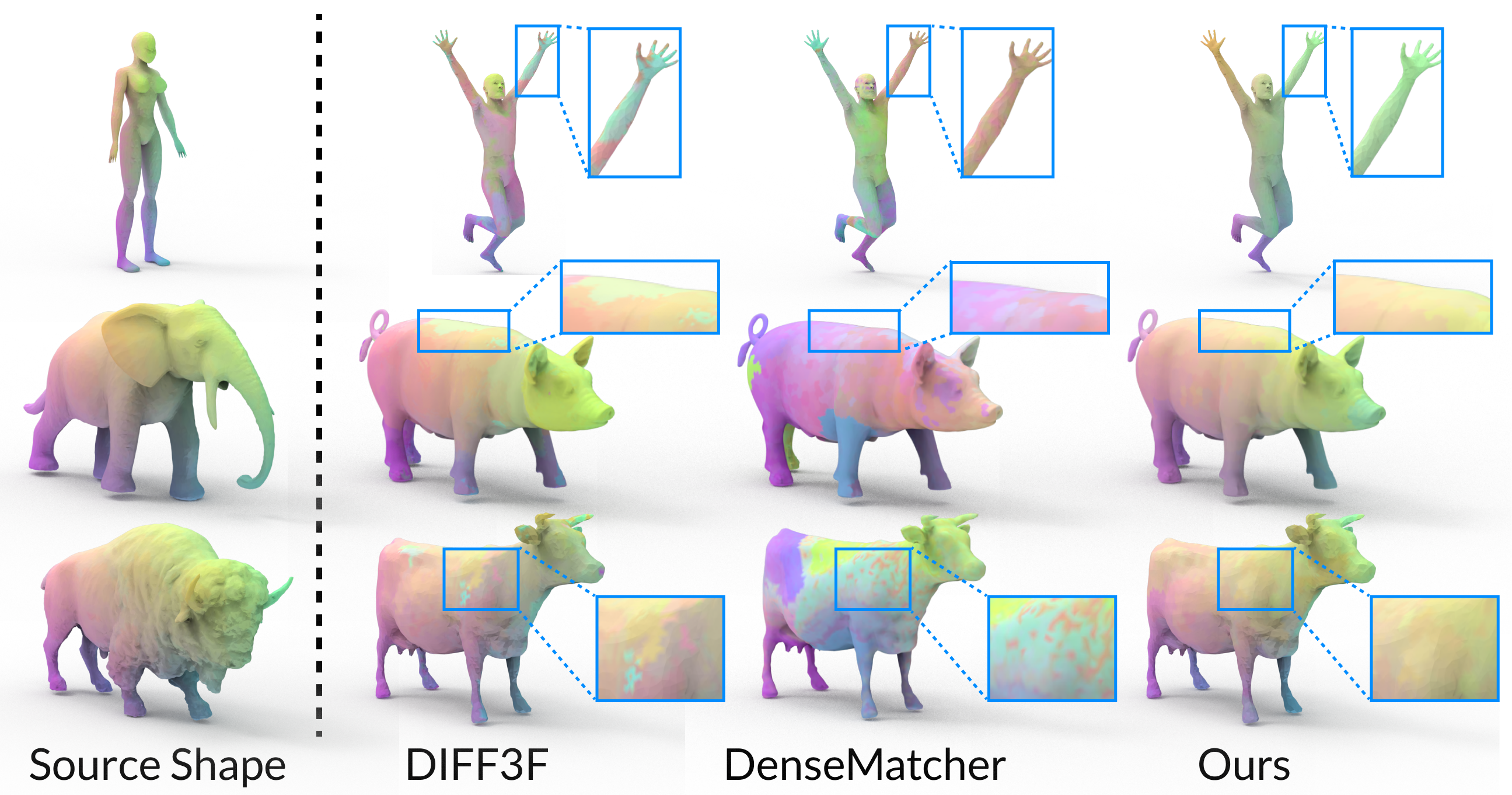}
    \vspace{-3mm}
    \caption{
    Visual comparison with the state-of-the-art methods DIFF3F~\cite{dutt2024diffusion} and DenseMatcher~\cite{zhu2024densematcher}. 
    Our method produces more accurate and coherent correspondences, while the results from DIFF3F and DenseMatcher are noisier and less consistent.
    }
    \label{fig: visual_cos_comp}
    \vspace{-4mm}
\end{figure}

\section{RESULTS AND EXPERIMENTS}
\subsection{Datasets and Implementation Details}  
Evaluations include six benchmark datasets: ShapeNet~\cite{chang2015shapenet}, ShapeNet Part~\cite{yi2016scalable}, Objaverse~\cite{deitke2023objaverse}, Objaverse(XL)~\cite{deitke2023objaversexl}, SHREC'19~\cite{melzi2019shrec}, and SHREC'20~\cite{dyke2020shrec}. 
We adapt our approach to three different pre-trained generative backbones: SDF-Diffusion~\cite{shim2023diffusion}, LAS-Diffusion~\cite {zheng2023locally}, and TRELLIS~\cite{xiang2024structured}. 
Regarding inference efficiency, we report the average runtime over 10 shape pairs: 
SDF-Diffusion, LAS-Diffusion, and TRELLIS take 1.6, 1.3, and 2.8 seconds per pair, respectively.
For SDF-Diffusion and LAS-Diffusion, the global semantic feature $\mathcal{F}_{G}$ is extracted from the 1st layer of the generator’s decoder, whereas the local geometric features $\mathcal{F}_{L}$ are obtained from the 2nd and 3rd layers. 
For TRELLIS, $\mathcal{F}_{G}$ is extracted from the 4th layer of the generator and the local geometric features $\mathcal{F}_{L}$ are taken from the 6th, 8th, and 10th layers.
The diffusion timestep $t$ is set to 45 for LAS-Diffusion, 50 for SDF-Diffusion, and 12 for TRELLIS.
%
%

%
\subsection{Experimental Settings}
To evaluate the effectiveness of our method, we compare it with various baselines on both shape correspondence and co-segmentation.
\paragraph{Shape correspondence.} 
We compare our approach with 
DPC~\cite{lang2021dpc}, SE-ORNet~\cite{deng2023se}, 3D-CODED~\cite{groueix20183d}, FM+WKS~\cite{ovsjanikov2012functional}, DenseMatcher~\cite{zhu2024densematcher}, and DIFF3F~\cite{dutt2024diffusion}. 
Following~\cite{dutt2024diffusion, deng2023se}, we evaluate 
on both SHREC'19~\cite{melzi2019shrec} (44 human scans with 430 annotated test pairs) and SHREC'20~\cite{dyke2020shrec} 
(featuring animals in diverse poses with expert-labeled non-isometric correspondences). 
We evaluate correspondence accuracy using the provided ground-truth annotations.
\vspace{-1mm}
\paragraph{Shape co-segmentation.} We compare our method with various baselines:
BAE-Net~\cite{chen2019bae}, RIM-Net~\cite{niu2022rim}, SATR~\cite{abdelreheem2023satr}, DAE-Net~\cite{chen2024dae}, and DenseMatcher~\cite{zhu2024densematcher}. 
Following~\cite{chen2019bae}, we evaluate shape 
on ShapeNet Part~\cite{yi2016scalable}, focusing on the following 13 categories: \texttt{airplane}, \texttt{cap}, \texttt{chair}, \texttt{earphone}, \texttt{guitar}, \texttt{knife}, \texttt{lamp}, \texttt{motorbike}, \texttt{mug}, \texttt{pistol}, \texttt{rocket}, \texttt{skateboard}, and \texttt{table}.
In our setting, co-segmentation is performed by transferring part labels from a labeled source shape to an unlabeled target shape. For each position on the target shape, we estimate its corresponding location on the source shape using our correspondence method, and then assign the part label from the source to the target.

\begin{figure}[t]
    \centering
    \includegraphics[width=0.99\linewidth]{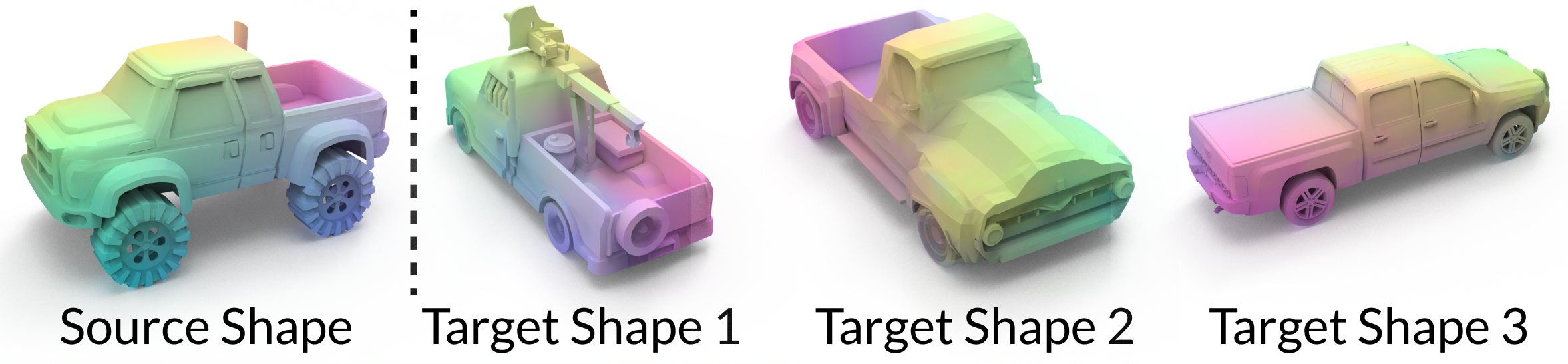}
    \vspace{-3mm}
    \caption{
    Visual correspondence results on shapes with different orientations. 
    By leveraging TRELLIS~\cite{xiang2024structured}, our method can produce meaningful correspondences even if the shapes are not canonically aligned.
    }
    \label{fig: rotation_invariance}
    \vspace{-3mm}
\end{figure}

\subsection{Quantitative Comparison}

\paragraph{Evaluation metrics.}
We evaluate correspondence estimation using two
standard metrics, following~\cite{dutt2024diffusion}: average correspondence error and correspondence accuracy.
Besides, we report two conventional metrics,~\ie, average geodesic distance and
edge distortion ratio, to assess the accuracy and smoothness of the correspondences.
For co-segmentation, we follow~\cite{chen2019bae} and report per-part IoU, averaged over all parts and shapes in each category. 
Please see Supplementary material Section A for details.

Table~\ref{tab: pointwise} reports the quantitative results on correspondence estimation. As SDF-Diffusion and LAS-Diffusion are not trained on human/animal shapes, we evaluate only the TRELLIS backbone on SHREC'19 and SHREC'20. Without additional training, our method outperforms prior approaches across all metrics, demonstrating its effectiveness.
Table~\ref{tab: geo_smooth} reports the quantitative results on geodesic accuracy error and smoothness estimation.  Here, we evaluate the TRELLIS backbone on SHREC'19 and SHREC'20. Again, even without additional training, our method surpasses prior approaches in both metrics.
Table~\ref{tab: Co_seg_IOU} presents the results on co-segmentation.
Our method with different backbones consistently outperforms state-of-the-art methods for all categories on all metrics, showing the robustness and generalizability of our framework design.
%

%
\subsection{Qualitative Comparison}
\paragraph{Shape correspondence}
We provide visual comparisons of our method with state-of-the-art approaches DIFF3F~\cite{dutt2024diffusion} and DenseMatcher~\cite{zhu2024densematcher}, which are the most relevant to ours---both methods establish 3D correspondences by utilizing 2D image features. 
Figure~\ref{fig: visual_cos_comp} shows qualitative comparisons with DIFF3F and DenseMatcher.
Our method yields more accurate correspondences on both human and animal shapes, whereas DIFF3F and DenseMatcher tend to 
produce noticeably noise in results.
This comparison reveals the domain gap between 2D and 3D representations, further showing the effectiveness of our 3D-native approach.
%
%

\begin{figure*}[t]
    \centering
    \includegraphics[width=\linewidth]{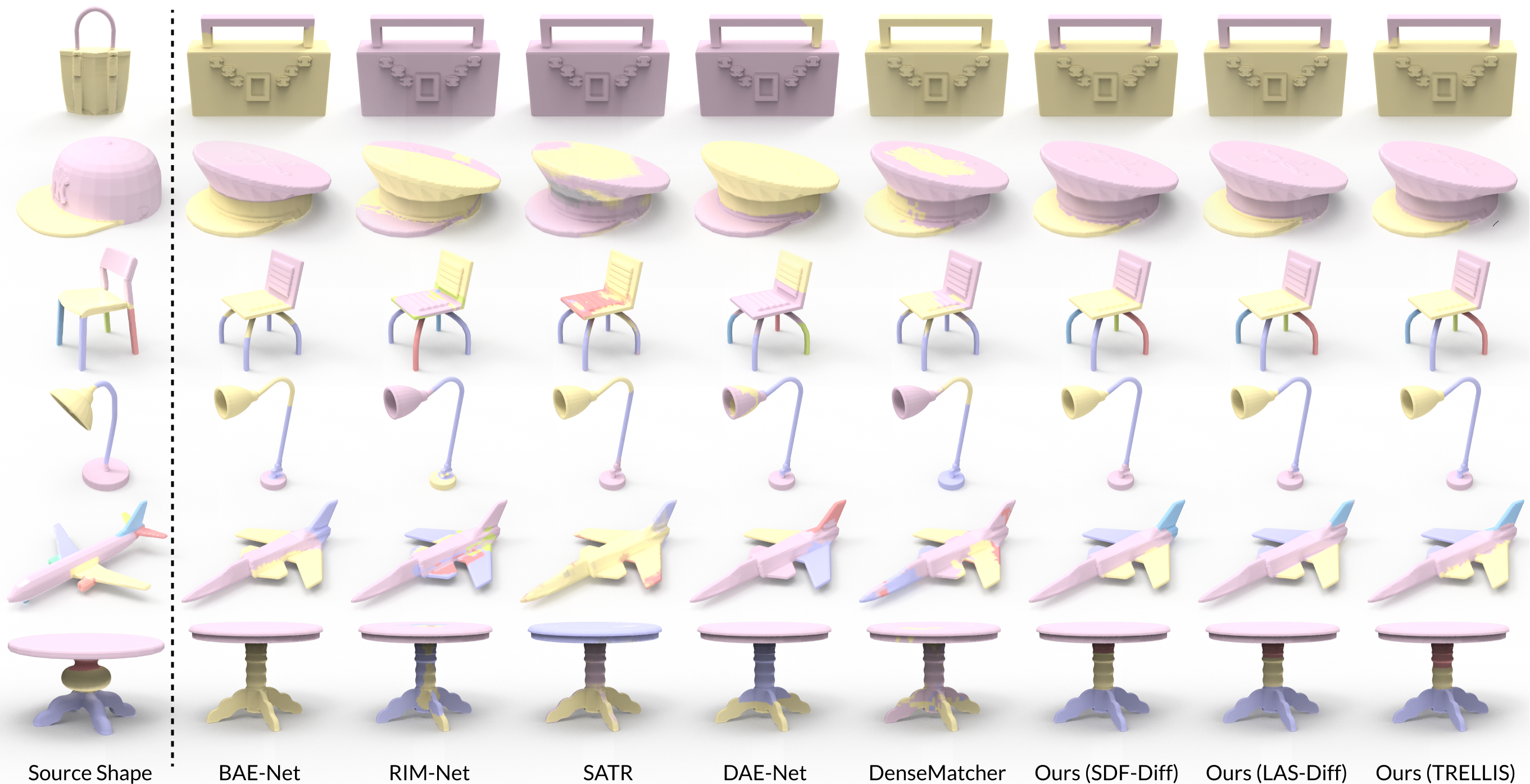}
    \caption{Visual comparisons between our approach and other state-of-the-art methods for shape co-segmentation, including BAE-Net~\cite{chen2019bae}, RIM-Net~\cite{niu2022rim}, SATR~\cite{abdelreheem2023satr}, DAE-NET~\cite{chen2024dae}, and DenseMatcher~\cite{zhu2024densematcher}. We apply our approach to three different baseline architectures, i.e., SDF-Diffusion~\cite{shim2023diffusion}, LAS-Diffusion~\cite{zheng2023locally}, and TRELLIS~\cite{xiang2024structured}, all of which achieve more accurate and consistent co-segmentation results than prior methods, especially in challenging regions such as label boundaries and part connections.
    }
    \label{fig: vis_co_seg}
\end{figure*}

%
\vspace{-2mm}
\paragraph{Shape co-segmentation}
Figure~\ref{fig: vis_co_seg} presents visual comparisons of our method with existing baselines on co-segmentation.
For all three 3D generative backbones, our framework consistently produces more precise parts co-segmentation with clear and well-aligned boundaries. 
As the third row of Figure~\ref{fig: vis_co_seg} shows, our method is able to co-segment all the four legs as distinct parts. In contrast, baseline methods tend to merge these substructures.
%


More visual comparison results on shape correspondence and co-segmentation are shown in Supplementary Sections B and C.

\vspace{-1mm}
\subsection{More Visual Results}
\label{ssec:visual_res}
\begin{figure*}[t]
    \centering
    \includegraphics[width=0.98\linewidth]{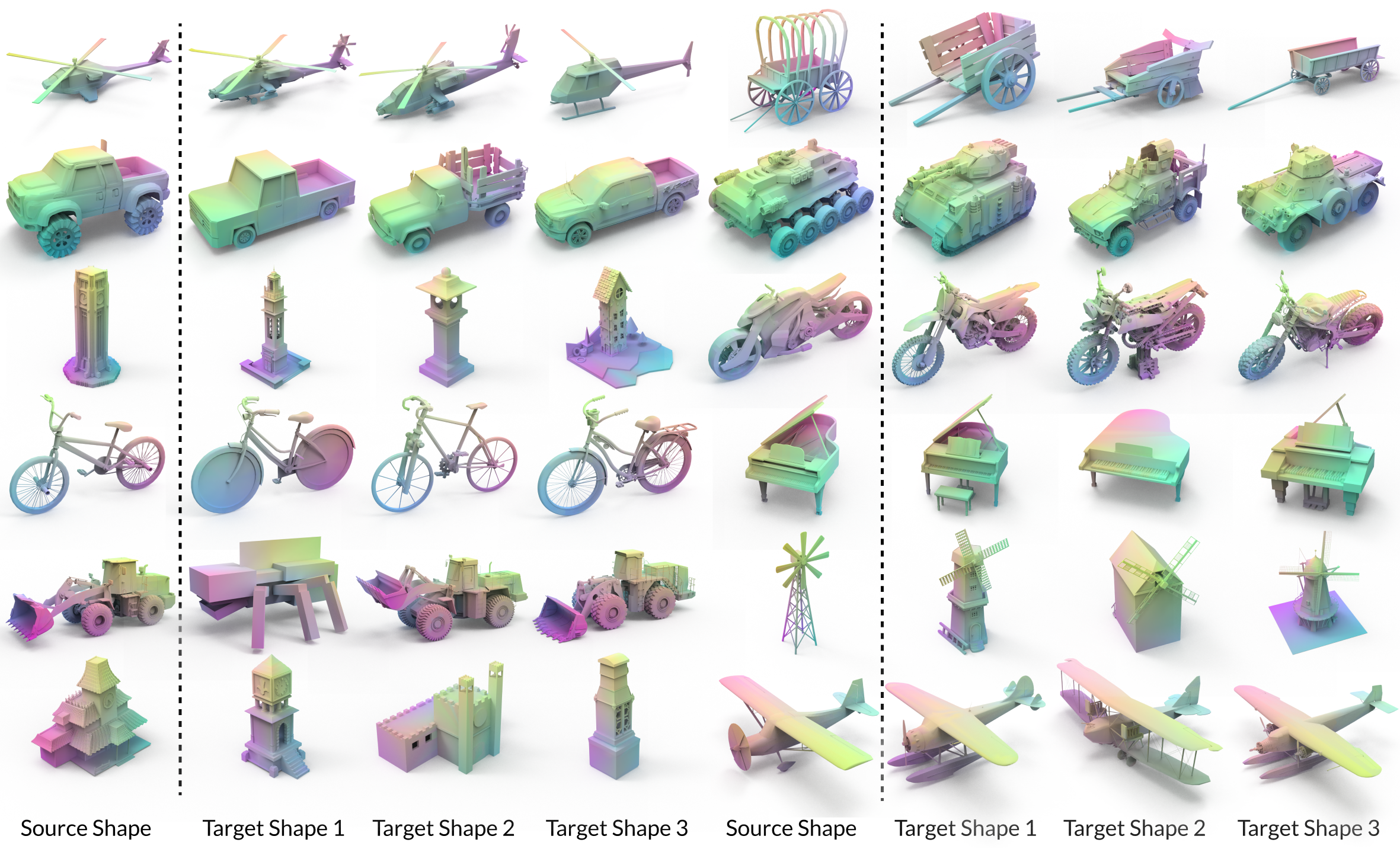}
    \caption{Visual results for 3D semantic correspondence. Our method effectively establishes semantic correspondences between shapes with varying geometries and topologies, and generalizes well across different object categories, demonstrating robustness to structural and semantic variations.}
    \label{fig: vis_dense}
\end{figure*}

\begin{figure*}[t]
    \centering
    \includegraphics[width=\linewidth]{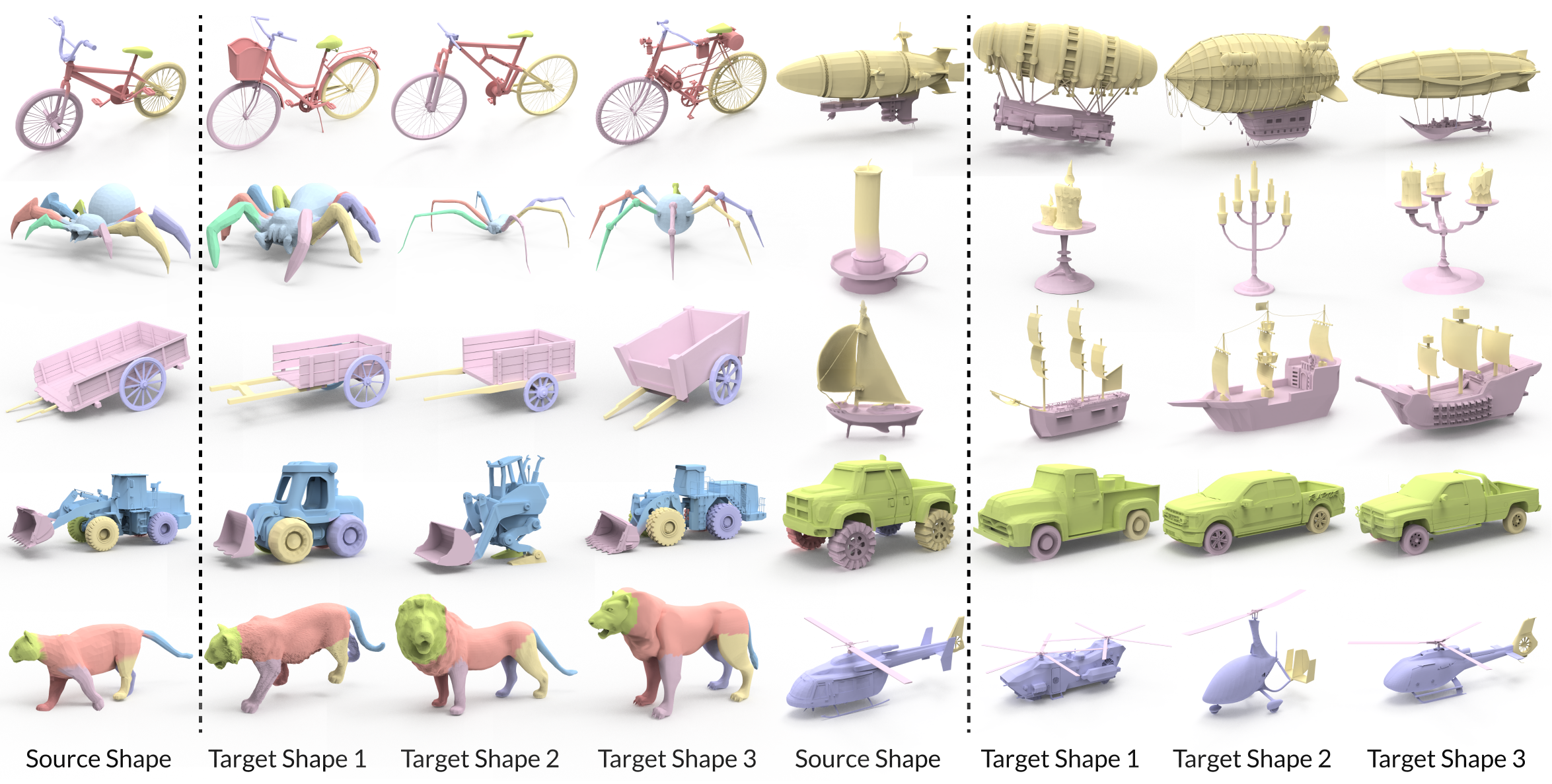}
    \caption{Visual results for shape co-segmentation. Our method delivers precise and semantically meaningful co-segmentation beyond simple spatial alignment. As shown in the second row (right), our method accurately co-segments the candle and holder, despite large variations in their relative positions.}
    \label{fig: vis_co_seg_obj}
\end{figure*}

Figures~\ref{fig: vis_dense} and~\ref{fig: vis_co_seg_obj} present additional qualitative results of correspondence and co-segmentation produced by our approach.
These results are generated using TRELLIS~\cite{xiang2024structured}, as it is trained on Objaverse and thus supports evaluation on more diverse shapes.

Further, since Objaverse contains unaligned shapes with varying orientations, TRELLIS learns to capture pose and orientation variations during training. 
Though our method does not explicitly consider 
rotation invariance, we observe that it appears to inherit
this property of TRELLIS to establish correspondences, even when the source and target shapes have different orientations; see Figure~\ref{fig: rotation_invariance}.
More visual results on shape correspondence and co-segmentation are shown in Sections D and E of the supplementary materials.

%

\subsection{Applications}

\begin{figure*}[t]
    \centering
    \includegraphics[width=0.98\linewidth]{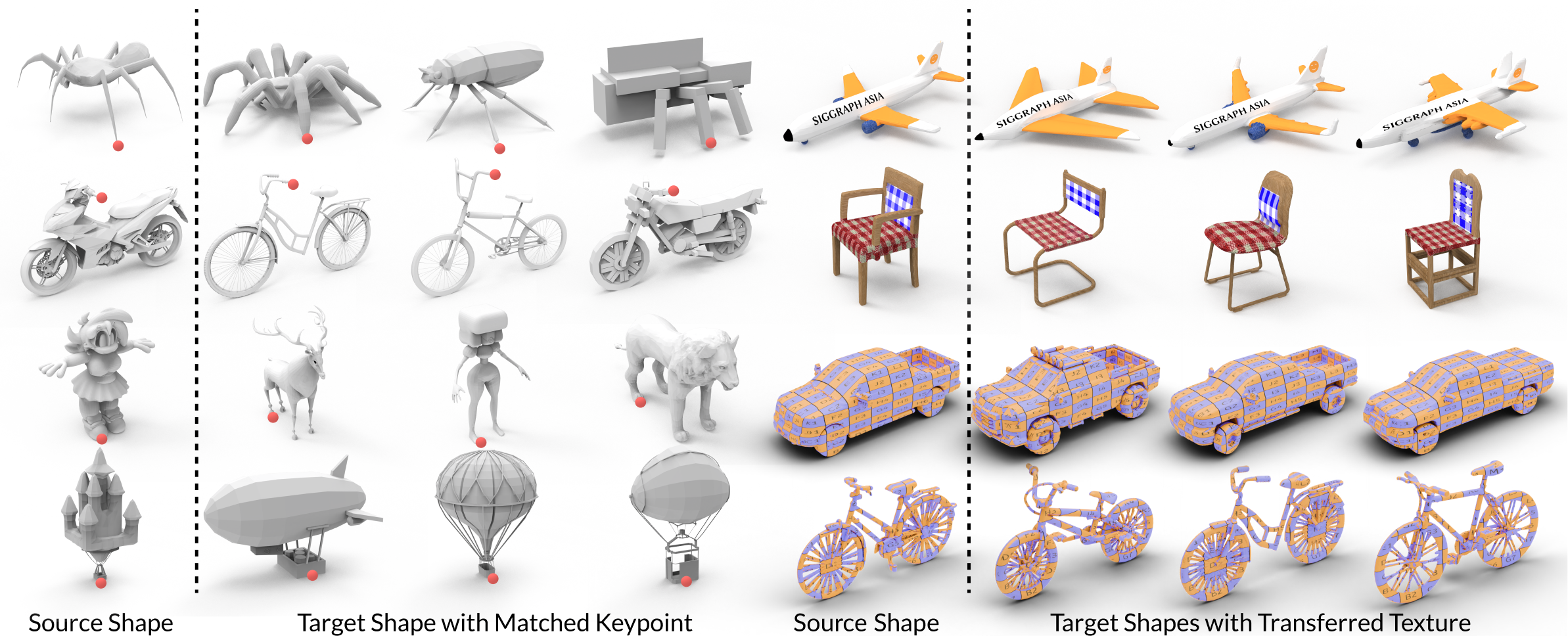}
    \caption{Visual results for keypoint matching and texture transfer. 
    The left part shows keypoint matching results, where our method reliably finds semantically corresponding positions across shapes, even in challenging cross-category scenarios, ~\eg, aligning a human foot to a deer’s foot (third row). The right part demonstrates texture transfer, where detailed patterns such as airplane text are accurately preserved and mapped onto target shapes, showcasing the effectiveness of our approach. On the third and fourth rows of the right panel, we present results on transferring high-frequency chessboard textures while highlighting patch boundaries and numbering each patch. As shown, our method preserves the chessboard structure even with high-frequency patterns.}
    \label{fig: keypoint}
\end{figure*}

Leveraging the estimated correspondences, our method enables faithful texture transfer and keypoint matching across diverse 3D shapes.
As Figure~\ref{fig: keypoint} (right) shows, texture patterns are preserved and aligned consistently with the corresponding semantic regions. To achieve this, our method leverages the part-wise UV maps available in Objaverse~\cite{deitke2023objaversexl}. Given a source mesh and a target mesh, we first establish semantic correspondences between their respective parts using our approach.
Then, for each matched part pair, we map the texture from the source part’s UV map to the UV coordinates of the corresponding target part.
This part-level transfer ensures semantic alignment and appearance consistency.

%
Figure~\ref{fig: keypoint} (left) shows our keypoint matching results. 
Given a user-specified keypoint on the source shape, our framework is able to automatically locate the semantically corresponding position on the target shapes via the estimated semantic correspondence.
The results remain consistent and accurate for diverse geometries and categories.
From the third row of Figure~\ref{fig: keypoint} (left), we can see that when a keypoint is placed on a human foot, our method can correctly transfer it to the corresponding position of a deer, demonstrating our method's robust cross-category semantic-matching capabilities.
See Supplementary Section F for more visual results.

%

%

%
\subsection{Ablation Studies}
\label{sec:ablation}
Due to page limit, see Supplementary Section G for ablation studies.

%
%

\section{CONCLUSION}
We presented a new framework, namely \name (\abb), for 3D semantic correspondence by leveraging features extracted from pre-trained 3D generative models. 
In summary, \name is a multiresolution representation composed of global semantic features and local geometric features, designed to capture both high-level structural context and fine-grained geometric details.
Further equipped with the progressive global-to-local matching strategy, our framework is able to construct accurate and semantically consistent correspondence, without requiring additional training.
Our approach demonstrates strong generalization across diverse shape categories and topologies, achieving state-of-the-art performance on standard benchmarks for correspondence estimation and shape co-segmentation. 
Moreover, we showcase its utility in real-world applications such as texture transfer and keypoint matching, highlighting its robustness and versatility.
%
%
%
\paragraph{Limitations and Future Work}
While our method shows strong performance in estimating semantic correspondences for a wide variety of 3D objects, its effectiveness inherently depends on the capability of the pre-trained generative model. If a target shape lies far outside the training distribution of the generative prior, the extracted features may not be able to encode meaningful semantics or geometric cues, leading to degraded correspondence performance. Although our approach demonstrates generalization to diverse categories using foundation models like TRELLIS, domain shifts or underrepresented categories can still pose challenges. 
A potential research direction is to enhance the robustness of generative models by incorporating additional fine-tuning process.

Moreover, our current framework is designed for static shape pairs.
While we demonstrate successful transfer across category and topology, the framework does not directly address correspondences in time-varying shape sequences. 
Extending our approach to support temporally consistent correspondence in dynamic 4D datasets would be a promising and impactful direction.
For instance, tracking the foot region across frames in a sequence, where both a human and a deer are running, requires maintaining semantic consistency over time, despite continuous deformation and pose variation.
Incorporating spatiotemporal priors could help capture such dynamics, paving the way for robust and temporally stable correspondence estimation in complex real-world scenarios.

%


\bibliographystyle{ACM-Reference-Format}
\bibliography{reference}


\end{document}